\documentclass{article}

\usepackage{iclr2024_conference,times}

\usepackage[utf8]{inputenc}
\usepackage[T1]{fontenc}
\usepackage[dvipsnames,table]{xcolor}
\usepackage[colorlinks=true,linkcolor=red,citecolor=blue,urlcolor=BrickRed]{hyperref}
\usepackage{url}
\usepackage{nicefrac}
\usepackage{microtype}
\usepackage{caption}
\usepackage{float}
\usepackage{subcaption}
\usepackage{enumitem}
\usepackage{pifont}
\usepackage{algorithm}
\usepackage{algpseudocode}
\usepackage{algorithmicx}
\usepackage{tabularx}
\usepackage{adjustbox}
\usepackage{multirow}
\usepackage{dsfont}
\usepackage{graphicx}
\usepackage{amsmath}
\usepackage{amssymb}
\usepackage{booktabs}

\title{Helicase: Uncertainty-Guided Supply Chain Knowledge Graph Construction with Autonomous Multi-Agent LLMs}

\author{Yunbo Long$^{a,*}$, Haolang Zhao$^{a}$, Ge Zheng$^{a}$, Alexandra Brintrup$^{a,b}$\\
$^{a}$Department of Engineering, University of Cambridge, Cambridge, UK\\
$^{b}$The Alan Turing Institute, London, UK\\
\texttt{yl892@cam.ac.uk}\\
{\small $^{*}$Corresponding author.}
}

\iclrfinalcopy

\begin{document}
\maketitle
\lhead{Preprint.}

\begin{abstract}
LLM-based multi-agent systems have been widely adopted for knowledge retrieval and report generation, synthesizing known information through web search and textual reasoning. However, many critical information tasks in supply chains are not simple one-shot queries: they are structural inference problems requiring multi-hop reasoning across complex, fragmented web resources. Questions such as \textit{``Which Tesla components use lithium from Australian mines?''} have no answer in any single document; answers must be computationally synthesized through the autonomous construction and analysis of dynamic knowledge graphs assembled from fragmented, heterogeneous sources. Moreover, such discovery processes must be uncertainty-aware: decisions depend not only on answers but on calibrated confidence in their reliability, traceable to source quality and reasoning consistency.
To address this capability gap, we propose \textit{Helicase}, an autonomous multi-agent LLM system for uncertainty-guided supply chain knowledge graph construction. \textit{Helicase} decomposes high-level supply-chain queries into executable investigation plans, coordinates specialized web-search, reasoning, and coding agents through iterative verification loops, and incrementally constructs query-specific supply chain knowledge graphs with per-fact uncertainty annotations. Its three-layer uncertainty framework tracks uncertainty at the action, trajectory, and memory layers, enabling both structural inference and calibrated confidence assessment.
To evaluate autonomous reasoning across the full complexity spectrum, we introduce SCQA (Supply Chain Query Assessment), a benchmark of 80 supply chain queries organized into four quadrants spanning single-hop to multi-hop inference under both high and low data visibility. Our experiments demonstrate that frontier LLMs and existing agent frameworks fail on SCQA: they cannot generate or reason over the network models required as intermediate artifacts, nor do they produce quantifiable uncertainty estimates. In contrast, Helicase successfully automates the end-to-end discovery process, substantially outperforming all baselines on answer accuracy across the four quadrants and is the only system producing calibrated uncertainty estimates over both discovered entities and their relationships. We further illustrate practical use through two case studies on real-world supply chains in the EV battery and consumer-chemical sectors.
This work establishes a new frontier for autonomous information discovery for supply chain research, shifting from passive, retrieval-augmented systems to proactive, uncertainty-guided, graph-based discovery. The code is available at \href{https://github.com/Yunbo-max/Helicase}{\textcolor{red}{https://github.com/Yunbo-max/Helicase}}.
\end{abstract}

\section{Introduction}\label{sec:introduction}

\begin{figure}[htbp]
    \centering
\includegraphics[width=\textwidth]{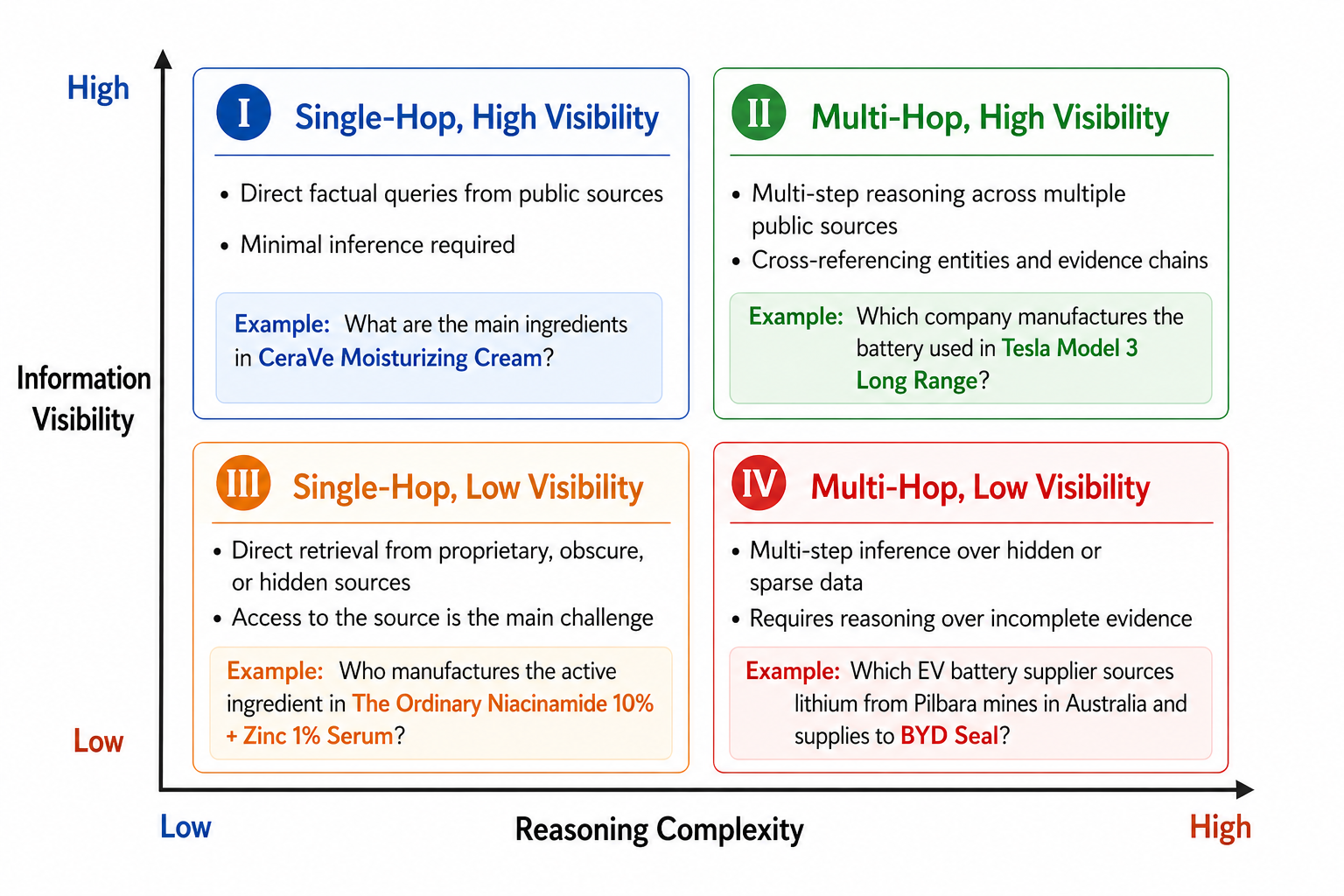}
    \caption{Overview of SCQA Research Benchmark Dataset Quadrants}
    \label{fig:scqa-quadrants}
\end{figure}



The advent of large language models (LLMs) like ChatGPT revolutionized how we access and synthesize information from the public resources. These models excel at answering common questions based on patterns in vast training data. However, pre-trained models possess no inherent, verifiable knowledge of proprietary information often sought in supply chain management, such as formulations or dynamic industrial relationships. Instead they can only deduce answers to queries from generic public text, making them unreliable for critical production research scenarios such as determining the exact formulation of a specific commercial product, tracing multi-tier ingredient dependencies, or assessing cross-industry dependencies and bottlenecks of products \citep{wu2025agentic}.

This gap is critical because questions about product composition lie at the heart of modern supply chain resilience and strategy. The ability to automatically answer complex queries about ingredient dependencies, for example \textit{"Which other products would be affected if the supply of a specific emulsifier fails?"} or  \textit{"What alternative materials could replace this polymer across our product lines without reformulation?"}, is essential for proactive risk management, ethical and sustainable sourcing, and strategic product development. Traditional supply chain intelligence systems are often constrained by their reliance on collecting distributed and fragmented product data from disparate sources. Their analysis is typically limited to constructing static product networks based on these incomplete datasets. Consequently, they fail to meet the demands of dynamic, real-world production environments, as they lack the formulation-level intelligence needed to map hidden vulnerabilities and opportunities across finished goods, categories, and companies \citep{huang2025deep}. Therefore, advancing beyond this static, data-limited paradigm represents a promising research venue in supply chain analytics \citep{zhang2025deep}.

Recently, the emergence of multi-agent deep research systems has promised a path toward generative knowledge discovery by actively searching, retrieving, and processing current web information \citep{xu2024multi}. However, existing multi-agent frameworks applied to supply chain or product research remain critically constrained by four fundamental limitations.
First, they lack true autonomous search capability. These systems either require pre-processed, clean text data \citep{jannelli2025agentic} or operate through rigid, predefined workflows and fixed prompts designed to retrieve only specific, known information types (e.g., supplier news) \citep{quan2024invagent}. Consequently, they cannot dynamically determine what to search for or how deep to investigate in response to ambiguous, complex questions. These systems thus act as retrieval tools rather than autonomous analysts, unable to handle diverse supply chain queries \citep{dong2025safesearch}.
A second critical flaw is the absence of verifiable, quantified uncertainty. Even state-of-the-art deep research agents powered by ChatGPT or Gemini produce narrative summaries with unverified citations prone to hallucination~\citep{huang2025deep}. More fundamentally, they lack mechanisms to assess source reliability. Nor can they quantify how uncertainty at each agent's output propagates through the multi-agent pipeline to affect the confidence of the final answer~\citep{nahar2025catch}. Consequently, while individual sources may be cited, the final recommendation itself carries no quantification of its uncertainty, a fatal shortcoming for supply chain decision-making, where managerial decisions must be based on evidence with explicit, traceable confidence levels \citep{tseng2018decision, simangunsong2012supply}.
In addition, the third flaw is the lack of explainability and auditability. The reasoning process of these systems is a black box. Tools like OpenAI's or Google's deep research agents cannot show how their multi-agent system arrived at a conclusion about a material bottleneck or a substitution pathway \citep{chan2024visibility}. This makes them unsuitable for strategic supply-chain planning, where understanding the reasoning behind a recommendation is as critical as the recommendation itself \citep{kosasih2024review}.
Finally, another challenge is agentic structural knowledge construction: supply chain research requires construction and inference on complex systems, not just narrative synthesis. This necessitates a multi-agent framework that integrates tools like coding agents to build structured intermediate models (e.g., knowledge graphs) and perform analysis. Existing deep research systems like OpenAI's or Google's, limited to web search and reasoning \citep{zhou2025efficient}, lack this capacity for structured analysis, making them inadequate for autonomous supply chain research.

\begin{figure}[htbp]
    \centering
\includegraphics[width=\textwidth]{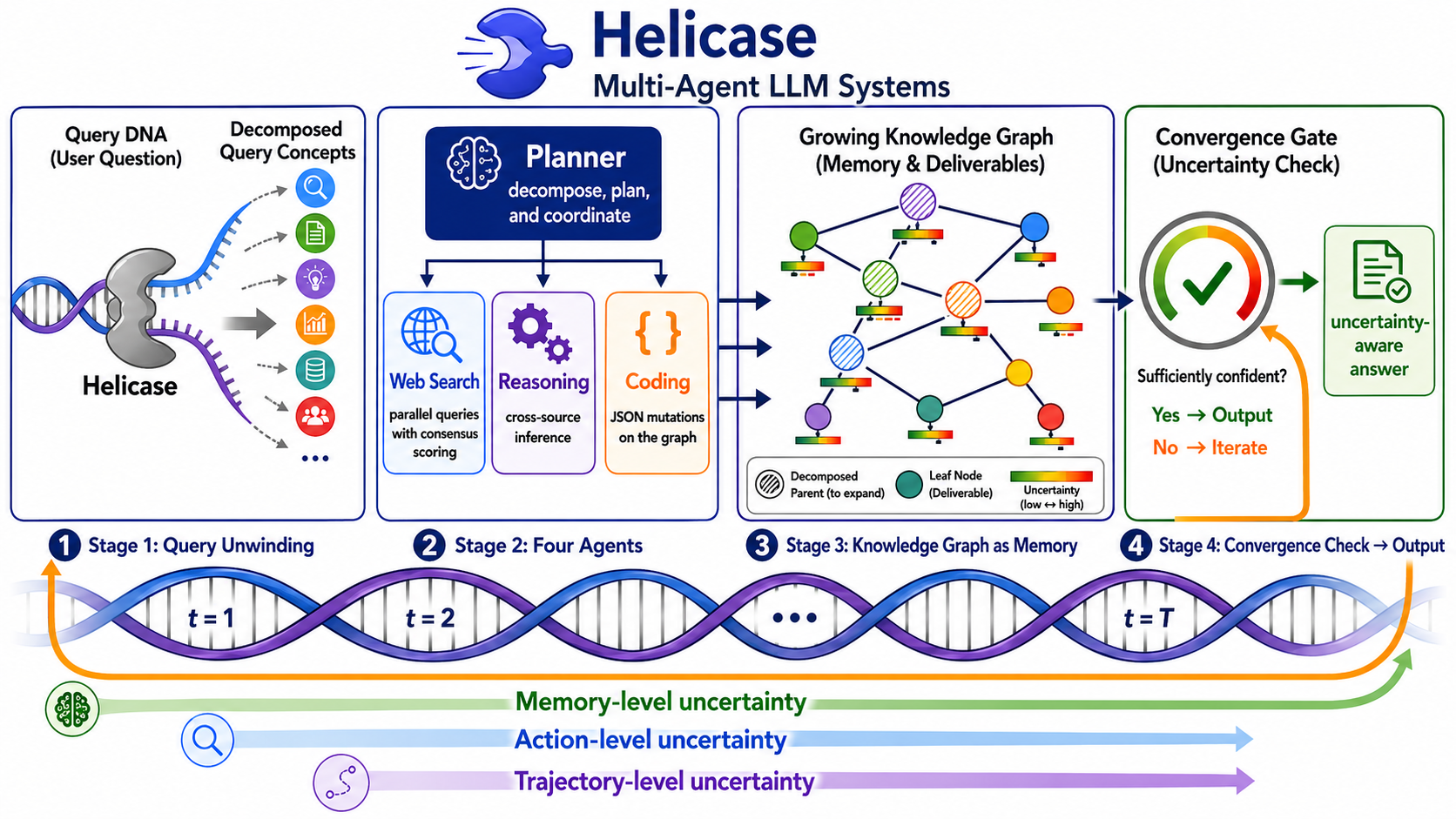}
    \caption{The Helicase system architecture: Query unwinding initiates a helical process of execution, mutation, repair, and convergence checking, all guided by uncertainty quantification.}
    \label{fig:helicase-architecture}
\end{figure}

To address these limitations, we propose \textit{Helicase}, an agentic LLM-based multi-agent system for autonomous supply chain discovery. The name evokes the biological helicase enzyme that unwinds DNA to expose the information encoded within; analogously, our system iteratively unwinds opaque supply chain queries, disentangling ambiguous terminology, latent dependencies, and fragmented evidence across heterogeneous web sources. Helicase introduces \emph{multimodal evidence} that supply chain investigations actually demand: regulatory filings and corporate sustainability reports as PDFs, ingredient and component disclosures embedded in product pages, tabular data in supplier directories and trade-press spreadsheets, and unstructured signals from interactive social platforms (e.g., Twitter/X, LinkedIn posts, industry forums) where supplier announcements, recall notices, and disruption alerts often surface before they are formally indexed. As shown in \autoref{fig:helicase-architecture}, Helicase implements a closed-loop helical process that integrates contextual query expansion, multi-source evidence harvesting with reliability scoring, and knowledge-graph construction via a coding agent. Crucially, it produces not a narrative summary but a validated, structured knowledge graph annotated with entities and their relationship uncertainty, directly usable for downstream supply chain analysis.
To guide our investigation, we formulate three research questions aligned with the agentic-SCM research agenda:
\begin{itemize}
\item \textbf{RQ1 (Capability):} How can autonomous AI systems answer supply chain questions whose answers are scattered across multiple sources and not recorded in any single document?
\item \textbf{RQ2 (Trustworthiness):} How can such systems produce verifiable, reliable answers that practitioners can confidently use for decision-making?
\item \textbf{RQ3 (Architecture):} How can autonomous supply chain discovery systems be designed to adapt to questions of varying difficulty and topic, while constructing reliable, high-quality supply chain knowledge?
\end{itemize}

To answer these questions, we introduce and evaluate Helicase on the Supply Chain Query Assessment (SCQA) benchmark, covering four complexity quadrants. Our results show that frontier LLMs and existing agentic frameworks fail on structural supply chain inference, while Helicase successfully automates end-to-end discovery with explicit uncertainty quantification. Our contributions are threefold: (1) the Helicase architecture for uncertainty-aware, structural supply-chain discovery, an agentic response to the coordination and trust challenges long identified in supply chain MAS research \citep{xu2021bots}; (2) a three-layer uncertainty quantification framework (action, trajectory, memory) addressing the hallucination and verification concerns raised by recent agentic-SCM studies \citep{jannelli2025agentic,menache2025generative}; and (3) the release of SCQA, the first benchmark designed to rigorously evaluate agentic supply chain discovery across four complexity quadrants.

\section{Related Work}

\label{sec:related_work}

\subsection{From Classical MAS to Agentic LLMs in Supply Chain Management}

Multi-agent systems (MAS) have been studied in supply chain management since the early 2000s for their potential in autonomous decision-making, but \citet{xu2021bots} documented that this research programme largely stalled in industrial practice due to four persistent barriers: (i) the high cost of designing and debugging bespoke agent logic, (ii) poor scalability beyond toy scenarios, (iii) fragile coordination between agents, and (iv) the inability of rule-based agents to generalise beyond their hand-crafted knowledge. Recent advances in agentic LLMs directly address these barriers: LLM-backed agents begin operation with broad general knowledge, communicate in natural language, use tools flexibly, and can be composed without specialist programming \citep{bubeck2023sparks, li2025chatsync}. Early applications to supply chain negotiation \citep{jannelli2025agentic} and logistics synchronisation \citep{li2025chatsync} illustrate the promise, while recent work cautions that agentic LLMs inherit managerial biases \citep{chen2025manager} and may produce confidently wrong decisions without explicit verification \citep{menache2025generative}. Our work sits at this intersection: we show that agentic LLMs can drive end-to-end supply chain discovery, but only when equipped with explicit, multi-layer uncertainty quantification to contain hallucinations and calibrate downstream trust.

\subsection{Visibility, Complexity, and the Discovery Problem}

Modern supply chains operate with structural visibility gaps: intelligence on supplier product portfolios, ingredient origins, and hidden dependencies lies scattered across siloed, unstructured sources \citep{agrawal2024supply, kalaiarasan2023supply}. Supply networks are considered complex adaptive systems \citep{choi2001supply} in which indirect, cross-tier dependencies emerge from interactions no single participant fully observes. This invisibility conceals cascading risks and sustainable alternatives, demanding digital supply chain surveillance that actively uncovers what static systems miss.
To address this visibility gap, the emergent field of Digital Supply Chain Surveillance (DSCS) advocates the systematic, technology-enabled monitoring and discovery of supply chain intelligence from heterogeneous external sources \citep{brintrup2024digital}. DSCS moves beyond traditional enterprise systems, which track known entities within known relationships, towards actively discovering unknown entities, undocumented connections, and emerging risks. This paradigm shift transforms supply chain analytics from passive reporting to active investigation.
However, existing DSCS approaches remain critically constrained by three interrelated limitations. First, they are passive \citep{zheng2025enhancing}, operating on fixed, pre-collected corpora rather than actively searching for new information to fill knowledge gaps (\autoref{related:1}). Second, they are limited to non-complex, high-visibility queries, unable to synthesize answers for multi-hop, low-visibility questions that require constructing multi-step structural and logic knowledge(\autoref{related:2}). Third, they are not uncertainty-aware, providing no quantification of source reliability or confidence in synthesized conclusions (\autoref{related:3}).

\subsection{AI for Uncovering Supply Chain Visibility}
\label{related:1}

Efforts to uncover hidden supply chain relationships have progressed through three technological stages. Early machine learning approaches focused on link prediction to infer missing relationships within partially populated knowledge graphs, assuming a static, pre-constructed graph structure \citep{brintrup2018predicting, kosasih2025towards}. Subsequent NLP-based methods automated the extraction of supplier and product relationships from unstructured text (news articles, regulatory filings, and safety datasheets), enabling the construction of supply networks from document corpora \citep{wichmann2020extracting}. However, these systems also remain passive: they operate on fixed, pre-collected corpora and cannot actively explore new information to fill knowledge gaps.
Most recently, LLM-based multi-agent frameworks have introduced autonomous web search and reasoning capabilities, demonstrating utility in building supply chain knowledge graph fr vaccine suppliers\citep{zheng2025enhancing}. However, these systems remain confined to informational retrieval, synthesizing pre-existing facts from documents or databases. They lack structural discovery: the ability to construct query-specific network representations from fragmented, cross-domain sources to answer questions whose answers are not recorded in any single source.
Helicase bridges this gap: it actively constructs knowledge graphs under uncertainty-guided agents, discovering hidden product-ingredient dependencies that prior systems cannot.

\subsection{Characterizing Supply Chain Queries: Visibility and Hop Complexity}

\label{related:2}

Supply chain information is scattered across the open sources with different structure, reliability, and accessibility. The ability to perceive risk, trace product origins, and verify sourcing critically depends on what information is visible and how that information is accessed \citep{ivanov2016supply,tiwari2024supply}. 
High-visibility information exists in external but structured, accessible sources: government open-data portals, commercial supplier directories, and public company filings. These sources are known and queryable: the data exists, the schema is defined, and the access mechanism is clear. The challenge is retrieval, not discovery.
Low-visibility information, by contrast, is dispersed across unstructured, non-standardized, and distributed external sources, such as technical datasheets, PDF safety reports, news articles, vendor websites, and specialist trade publications. In such cases, relevant evidence may exist in the public domain, but its location, format, terminology, and source reliability are not known a priori \citep{kosasih2024towards}. Retrieving and validating this evidence is therefore difficult, often requiring iterative search, entity resolution, and cross-source verification. Single-hop queries require one discrete search action (a single API call, database query, or web search) to obtain an answer. Multi-hop queries require sequential, interdependent search steps, where the output of one search informs the next query, tracing a path across multiple sources \citep{chen2019understanding}.
For high-visibility and single-hop questions, research in supply chain risk management has developed manual and semi-automated methods to retrieve answers from structured databases and pre-collected document corpora \citep{brintrup2024digital}. However, existing DSCS approaches remain fundamentally passive: they rely on fixed sources and predefined schemas, and therefore cannot autonomously identify relevant sources not specified a priori, generate adaptive search strategies, or traverse multi-hop evidence chains across heterogeneous external data.
Consequently, they are unable to address multi-hop, low-visibility questions, precisely where the most critical risks and opportunities lie \citep{brintrup2024digital, kosasih2022machine}.

\subsection{Uncertainty Quantification for Supply Chain}
\label{related:3}

Uncertainty quantification is critical for trustworthy decisions in supply chain management, where data is frequently inaccurate, incomplete, or conflicting \citep{zhao2025uncertainty, simangunsong2012supply}. In the context of knowledge graph-based supply chain analytics, recent work has explored uncertainty-aware link prediction and fact confidence scoring for static, pre-constructed graphs \citep{aziz2021data}. These methods assign uncertainty scores to individual entities or relationships based on source quality or inference confidence, but they assume the graph is already built and fixed.
However, DSCS introduces a fundamentally different and more complex uncertainty challenge. Here, uncertainty is not static; it is dynamic and compositional. It arises not from a single model prediction, but from an end-to-end, multi-agent investigation and graph-construction process that actively builds query-specific supply chain knowledge structures in response to previously unseen questions. This process introduces multiple, interdependent sources of uncertainty, including source reliability (e.g., regulatory databases versus vendor blog posts), information conflicts across documents, entity-resolution errors (e.g., INCI name versus chemical name versus trade name), and the \textit{propagation} of errors through sequential, multi-hop reasoning steps \citep{kirchhof2025position, han2024towards}. Critically, these uncertainties are not independent: they interact and accumulate as evidence is retrieved, linked, and encoded into the knowledge graph.
Existing methods for LLM confidence calibration \citep{zhao2024saup} or static KG uncertainty \citep{aziz2021data} cannot capture this dynamic, propagating uncertainty. They treat each fact or prediction in isolation, offering no mechanism to track how uncertainty accumulates along an investigative trajectory or consolidates into the final answer presented to a decision-maker. To address this gap, we introduce an LLM-based consensus mechanism that quantifies compositional, propagating uncertainty through multi-agent deliberation, a challenge unresolvable by semantic embeddings or conventional NLP confidence measures.

\section{Helicase Multi-agent System}
\label{sec:methodology}

Open-ended supply chain questions pose three intertwined challenges that are difficult for a single LLM call to address reliably: locating evidence fragmented across heterogeneous public sources, integrating that evidence into an auditable structural representation of the supply chain, and tracking uncertainty so that well-supported facts can be distinguished from weakly supported inferences. \textit{Helicase} addresses these challenges through agent specialization and uncertainty-guided iterative knowledge graph construction. A \emph{planner agent} sits at the top of the hierarchy: it decomposes the user query into an initial action set and, at every subsequent iteration, generates new actions targeted at high-uncertainty regions of the current knowledge graph. Three types of agents execute these actions in parallel. \emph{Web search agents} perform multi-query, multi-source evidence retrieval and extraction from heterogeneous sources, including static HTML pages, PDF reports (e.g., regulatory filings, corporate sustainability reports, and risk disclosures), tabular records (e.g., CSV/Excel supplier directories and trade-press rankings), and publicly accessible signals from platforms such as Twitter/X, LinkedIn, and industry forums, where supplier announcements and disruption alerts may appear before they are formally indexed. Each modality is normalised by a dedicated reader (PDF text extraction with table preservation, CSV-to-record parsing, JS-rendered page fallback for interactive sites) before evidence enters the consensus layer. The web search agent then scores evidence quality through LLM-based consensus across $n$ parallel queries. \emph{Reasoning agents} synthesise the harvested evidence, perform cross-source inference, and identify which structural updates the new findings warrant. \emph{Coding agents} translate these decisions into deterministic, auditable JSON mutations on the knowledge graph by creating, merging, or revising nodes and edges with fuzzy entity matching (containment with LLM fallback) to prevent duplicates. The planner and worker agents are bound together by a helical loop: at each cycle, the current knowledge graph and its uncertainty landscape feed back to the planner, driving the next round of action generation, evidence harvesting, and graph enrichment, until uncertainty stagnation signals convergence. The evolving knowledge graph itself serves as both a structured memory of the investigation and the container for final, uncertainty-aware answers.

The system operates through helical iterations where each complete rotation progresses through three sequential uncertainty quantification stages: (1) evaluating individual action uncertainties via LLM-based factual consensus scoring as they execute, (2) computing trajectory-layer cross-iteration redundancy via embedding similarity against all prior actions, and (3) updating memory-layer uncertainty in the knowledge graph via multiplicative accumulation. This three-stage quantification creates a feedback loop where uncertainty measurements from one iteration inform action selection and strategy adaptation in the next. The system state at iteration $t$ is defined as $\langle \mathcal{A}^{(t)}, \mathcal{G}^{(t)}, U^{(t)} \rangle$, where $\mathcal{A}^{(t)}$ represents the action set, $\mathcal{G}^{(t)}$ the knowledge graph, and $U^{(t)} = (U_{\text{action}}^{(t)}, U_{\text{trajectory}}^{(t)}, U_{\text{memory}}^{(t)})$ the three-layer uncertainty state.

\subsection{Query Enrichment and Contextual Refinement}
\label{subsec:query-enrichment}

The query enrichment process begins with an initial decomposition of the input query $Q$ into an action set $\mathcal{A}^{(0)} = \{(a_i, g_i)\}_{i=1}^{k}$, where each action $a_i$ is a search or reasoning task and $g_i \in \{\texttt{web\_search}, \texttt{reasoning}, \texttt{coding}\}$ is its assigned agent type. The core innovation lies in the contextual action generation function $\Phi_{\text{gen}}$, a reasoning-driven process that dynamically extends $\mathcal{A}^{(t)}$ based on execution outcomes and updated knowledge in $\mathcal{G}^{(t)}$. Formally:
\begin{equation}
\mathcal{A}^{(t+1)} = \mathcal{A}^{(t)} \cup \Phi_{\text{gen}}\left( \mathcal{R}^{(t)}, \mathcal{G}^{(t)}, Q, U^{(t)} \right)
\end{equation}
where $\mathcal{R}^{(t)} = \{r_i^{(t)}\}$ denotes execution results from iteration $t$ and $U^{(t)}$ is the current three-layer uncertainty state. The function $\Phi_{\text{gen}}$ is implemented as a planner agent that receives: (1) execution results with per-action uncertainties $\{U_{\text{action}}(a_i)\}$, (2) the knowledge graph summary with high-uncertainty nodes highlighted, (3) the original query $Q$, and (4) per-concept search history with redundancy flags. It outputs targeted actions ranked by estimated uncertainty reduction (Section~\ref{subsec:execution-convergence}).

The enrichment process operates through iterative refinement: at each cycle, the reasoning agent analyzes accumulated evidence and uncertainty patterns to formulate more precise, targeted questions. This creates a closed-loop exploration pattern where questions evolve based on discovered evidence and uncertainty measurements, allowing the system to dynamically refocus investigation toward regions with the highest uncertainty reduction potential.

\subsection{Three-Layer Uncertainty Quantification}
\label{subsec:uncertainty-quantification}

Uncertainty in Helicase is quantified at three complementary scales, computed at the conclusion of each iteration $t$:
\begin{itemize}
\item \textbf{Action-layer:} measures the reliability of each individual agent execution (did the web search produce consistent results?).
\item \textbf{Trajectory-layer:} detects whether the system is still discovering new information or repeating prior work.
\item \textbf{Memory-layer:} consolidates all per-fact uncertainties stored in the knowledge graph $\mathcal{G}^{(t)}$ into a single system-wide confidence measure $U_{\text{memory}}^{(t)} \in [0,1]$, where 0 indicates full confidence and 1 indicates complete uncertainty.
\end{itemize}
Together, these layers create a feedback loop: uncertainty measurements from one iteration guide action selection in the next.

\paragraph{Action-layer Uncertainty Calculation}
Action-layer uncertainty $U_{\text{action}}(a_i)$ is computed differently for each agent type. For web search agents, $n$ independent queries return answer sets $\{A_1, \dots, A_n\}$, which are evaluated via \textit{LLM-based factual consensus scoring}. The LLM receives all search results and directly assesses factual agreement on a calibrated $[0, 1]$ scale:
\begin{equation}
U_{\text{action}}(a_i) = \text{LLM}_{\text{consensus}}(A_1, \dots, A_n \mid a_i)
\end{equation}
where the LLM evaluates whether the results report identical facts, entities, and quantities (low $U$), partially overlapping information with gaps (moderate $U$), or contradictory claims (high $U$). This approach directly measures factual consistency rather than lexical similarity, avoiding the failure mode of embedding-based dispersion where cosine similarity of paragraph-length text consistently returns near-unity values regardless of factual agreement, conflating stylistic variation with genuine disagreement.

The number of parallel queries $n \in [n_{\min}, n_{\max}]$ is dynamically determined by an LLM that assesses question difficulty, current KG state, and the target concept's uncertainty $U(v_{\text{target}})$. When $U(v_{\text{target}})$ is below a low-uncertainty threshold $\tau_{\text{low}}$, the system defaults to $n_{\min}$ to conserve resources; for high-uncertainty concepts, the LLM allocates up to $n_{\max}$ parallel queries to maximize evidence diversity.

For coding agents executing deterministic structured extraction, uncertainty is binary $U \in \{0,1\}$ based solely on extraction success or failure. For reasoning agents, synthesized conclusions are similarly evaluated via LLM consensus scoring, assessing internal consistency of the reasoning output.

This agent-specific approach ensures uncertainty quantification aligns with each agent's operational characteristics. The LLM consensus approach naturally handles the spectrum from complete agreement to outright contradiction, and critically distinguishes between genuine factual disagreement and mere differences in phrasing or emphasis.

\paragraph{Trajectory-layer Uncertainty: Cross-Iteration Redundancy}
After all actions in iteration $t$ are complete, the system computes trajectory-layer uncertainty $U_{\text{trajectory}}^{(t)}$ as a redundancy signal: the degree to which the current iteration repeats previously executed investigations rather than exploring additional evidence chains or unresolved aspects of the query. Let the action set at iteration $t$ be
\[
\mathcal{A}^{(t)}=\{a_1^{(t)}, a_2^{(t)}, \dots, a_{m_t}^{(t)}\},
\]
where $m_t=|\mathcal{A}^{(t)}|$. The set of all actions executed before iteration $t$ is
\[
\mathcal{A}^{(<t)}=\bigcup_{\tau=1}^{t-1}\mathcal{A}^{(\tau)}.
\]

For each current action $a_k^{(t)} \in \mathcal{A}^{(t)}$, we compute its maximum semantic similarity to any previously executed action:
\begin{equation}
\rho_k^{(t)}
=
\max_{a_j^{(\tau)} \in \mathcal{A}^{(<t)}}
\text{sim}\left(
\text{embed}(a_k^{(t)}),
\text{embed}(a_j^{(\tau)})
\right),
\quad \tau < t .
\end{equation}
Here, $a_k^{(t)}$ denotes the $k$-th action in the current iteration, while $a_j^{(\tau)}$ denotes the $j$-th action from a previous iteration $\tau$. The function $\text{embed}(\cdot)$ maps each action description to a semantic vector; implementation details of the embedding model are provided in the experimental setup.
The trajectory-layer uncertainty is then computed as the average redundancy of the current action set:
\begin{equation}
U_{\text{trajectory}}^{(t)}
=
\frac{1}{m_t}
\sum_{k=1}^{m_t}
\rho_k^{(t)} .
\end{equation}

This yields low $U_{\text{trajectory}}^{(t)}$ when current actions target previously unexplored evidence chains or unresolved aspects of the query, and high $U_{\text{trajectory}}^{(t)}$ when they are semantically close to prior actions. In this sense, trajectory-layer uncertainty captures diminishing returns in the investigation: a high value suggests that the system is revisiting similar search or reasoning directions, prompting the planner $\Phi_{\text{gen}}$ to pivot toward independent lines of inquiry or terminate if memory-level uncertainty has also stagnated.

\paragraph{Memory-layer Uncertainty and Multiplicative Accumulation}
Memory-layer uncertainty $U_{\text{memory}}^{(t)}$ is computed by consolidating fact-level uncertainties stored in the knowledge graph $\mathcal{G}^{(t)}$. Each node and edge in the KG is treated as a fact $f$ with an associated uncertainty score $U^{(t)}(f)$. Let $\mathcal{E}_f^{(t)} \subseteq \mathcal{A}^{(t)}$ denote the set of actions in iteration $t$ that provide supporting evidence for fact $f$. For an existing fact, uncertainty is updated through multiplicative accumulation:
\begin{equation}
U^{(t)}(f)
=
U^{(t-1)}(f)
\prod_{a_k^{(t)} \in \mathcal{E}_f^{(t)}}
U_{\text{action}}(a_k^{(t)}),
\end{equation}
where $U_{\text{action}}(a_k^{(t)})$ is the action-layer uncertainty of the $k$-th action in iteration $t$ that produced evidence for fact $f$. If no new evidence is found for $f$ in iteration $t$, then $\mathcal{E}_f^{(t)}=\emptyset$ and $U^{(t)}(f)=U^{(t-1)}(f)$.

The multiplicative update reflects \textit{compounding confirmation}: assuming that supporting investigations provide sufficiently independent evidence, each consensus-bearing action with $U_{\text{action}}(a_k^{(t)}) < 1$ reduces the fact's uncertainty multiplicatively. In contrast, a minimum-based update,
\[
U^{(t)}(f)=\min\left(U^{(t-1)}(f), U_{\text{action}}(a_k^{(t)})\right),
\]
records only the most confident single observation and cannot represent cumulative corroboration. When repeated searches return similar moderate uncertainty values, the fact uncertainty remains unchanged rather than decreasing with additional supporting evidence.

For newly created nodes or edges, the fact has no prior uncertainty $U^{(t-1)}(f)$. We therefore initialize its uncertainty using the strongest evidence that introduced it:
\begin{equation}
U^{(t)}(f_{\text{new}})
=
\min_{a_k^{(t)} \in \mathcal{E}_{f_{\text{new}}}^{(t)}}
U_{\text{action}}(a_k^{(t)}).
\end{equation}
This initialization rule ensures that new entities inherit the uncertainty of the evidence that discovered them, rather than the residual uncertainty of the broader target concept. Subsequent investigations of the same fact are then accumulated multiplicatively, allowing well-confirmed facts to converge toward low uncertainty while disputed or weakly supported facts remain at elevated uncertainty.

The graph-wide memory-layer uncertainty is:
\begin{equation}
U_{\text{memory}}^{(t)}
=
\frac{1}{|V_{\text{active}}^{(t)}|}
\sum_{v \in V_{\text{active}}^{(t)}}
\max_{f \in F^{(t)}(v)}
U^{(t)}(f),
\end{equation}
where $V_{\text{active}}^{(t)}$ is the set of non-decomposed leaf nodes at iteration $t$, $F^{(t)}(v)$ is the set of facts involving entity $v$, and $U^{(t)}(f)$ is the current uncertainty of fact $f$ after all updates in iteration $t$. Decomposed parent concepts are excluded because they serve as structural groupings rather than direct search targets.

\subsection{Iterative Execution and Convergence}
\label{subsec:execution-convergence}

The system follows an uncertainty-driven control policy in which candidate actions are selected according to their estimated potential to reduce memory-layer uncertainty. After iteration $t$, the action planner $\Phi_{\text{gen}}$ analyzes the current knowledge graph $\mathcal{G}^{(t)}$ and uncertainty state $U^{(t)}$ to generate a candidate action set for the next iteration:
\[
\mathcal{C}^{(t+1)}
=
\{\tilde{a}_1^{(t+1)}, \tilde{a}_2^{(t+1)}, \dots, \tilde{a}_{q_{t+1}}^{(t+1)}\},
\]
where $\tilde{a}_\ell^{(t+1)}$ denotes the $\ell$-th candidate action proposed for iteration $t+1$.

The planner receives: (1) the full KG summary with per-node and per-edge uncertainty, (2) the set of high-uncertainty facts,
\[
\mathcal{F}_{\text{high}}^{(t)}
=
\{f \in \mathcal{F}^{(t)} : U^{(t)}(f) > \tau_{\text{high}}\},
\]
(3) per-concept search histories with redundancy flags, and (4) the trajectory-layer redundancy signal from the previous iteration. Each candidate action is ranked by estimated uncertainty reduction per unit cost:
\begin{equation}
\text{priority}\!\left(\tilde{a}_\ell^{(t+1)}\right)
=
\frac{
\Delta \hat{U}\!\left(\tilde{a}_\ell^{(t+1)}\right)
}{
\text{cost}\!\left(\tilde{a}_\ell^{(t+1)}\right)
},
\end{equation}
where
\begin{equation}
\Delta \hat{U}\!\left(\tilde{a}_\ell^{(t+1)}\right)
=
\max_{f \in \mathcal{F}_{\text{high}}^{(t)}}
\text{sim}
\left(
\text{embed}\!\left(\tilde{a}_\ell^{(t+1)}\right),
\text{embed}\!\left(s(f)\right)
\right)
\cdot
U_{\text{memory}}^{(t)}
\cdot
\alpha .
\end{equation}
Here, $s(f)$ denotes a textual representation of fact $f$, such as an entity label or relation triple; $\text{sim}(\cdot,\cdot)$ is cosine similarity in the embedding space; $\alpha$ is a scaling factor; and $\text{cost}(\tilde{a}_\ell^{(t+1)})$ is agent-type-dependent. The highest-priority candidates are selected to form the executed action set $\mathcal{A}^{(t+1)}$.
This prioritization focuses computational resources on actions that are semantically aligned with the most uncertain parts of the current knowledge graph. The planner also uses per-concept search histories to identify which prior actions produced low uncertainty, which remained inconclusive, and which were flagged as redundant because their trajectory similarity exceeded $\tau_{\rho}$. This allows the planner to avoid repeating low-yield search directions and to adapt its strategy for concepts where multiple evidence-gathering attempts remain inconclusive.

\paragraph{Convergence via Stagnation Detection}
The system terminates when memory-layer uncertainty exhibits \textit{sustained stagnation}: the relative change in $U_{\text{memory}}$ falls below a convergence threshold for multiple consecutive iterations, indicating that the investigation has exhausted its ability to reduce uncertainty through further search:
\begin{equation}
\text{converged} \iff \left|\frac{U_{\text{memory}}^{(t)} - U_{\text{memory}}^{(t-1)}}{U_{\text{memory}}^{(t-1)}}\right| < \delta_{\text{conv}} \text{ for } P \text{ consecutive iterations} \;\;\wedge\;\; t > 2
\end{equation}
where $\delta_{\text{conv}}$ is the relative change threshold and $P$ is the patience parameter requiring consecutive stagnant iterations. This approach is more robust than fixed absolute thresholds: it avoids premature termination from lucky initial consensus on incomplete information by requiring \textit{sustained} evidence of diminishing returns. The $t > 2$ guard ensures at least three full helix rotations complete before convergence is possible. A hard cap of $T_{\max}$ iterations prevents unbounded execution. The relative formulation also adapts naturally to different query complexities: queries that converge to different asymptotic uncertainty layers both trigger convergence when further iterations fail to meaningfully reduce uncertainty.

\section{SCQA Dataset}
\label{sec:dataset}

To rigorously evaluate autonomous systems across the full spectrum of supply chain discovery tasks, we introduce the \textbf{Supply Chain Query Assessment (SCQA)} benchmark. SCQA organises 80 queries spanning personal care, food and beverages, and electronics/automotive into four quadrants defined by two orthogonal dimensions.
Reasoning complexity distinguishes \textit{single-hop} queries (answerable from a single source document or database lookup) from \textit{multi-hop} queries that require chaining facts across multiple independent sources to synthesise an answer that exists in no single document. Information visibility distinguishes \textit{high-visibility} information (publicly documented in official product pages, regulatory filings, or well-known industry databases) from \textit{low-visibility} information that is fragmented across niche sources, not systematically recorded, or requires domain expertise to locate (e.g., upstream supplier identities, contract manufacturing relationships, intermediary procurement networks). We note that SCQA targets \textit{low-visibility} information, that is, data that exists somewhere in public sources but is difficult to find, requires cross-source reasoning to piece together, and often demands inference from indirect evidence (e.g., inferring a supplier relationship from co-occurrence in a sanctions filing), rather than \textit{invisible} information that is entirely proprietary or classified and cannot be discovered through open-source investigation.

These two dimensions yield four quadrants (20 queries each), illustrated in \autoref{fig:scqa-quadrants}:
\begin{itemize}
\item \textbf{Q1 (Single-hop, High-visibility)}: 20 queries; a single, well-documented source suffices. \textit{Example: ``Does Pantene Pro-V shampoo contain Sodium Lauryl Sulfate?''} (answerable from the product's official ingredient list.)
\item \textbf{Q2 (Multi-hop, High-visibility)}: 20 queries; multiple well-documented sources must be aggregated and filtered. Each individual fact is public, but the answer requires combining them. \textit{Example: ``What Johnson \& Johnson baby products are made in the USA and contain aloe vera?''} (requires cross-referencing manufacturing locations with ingredient lists across products.)
\item \textbf{Q3 (Single-hop, Low-visibility)}: 20 queries; the answer involves a single fact that is not prominently published, appearing only in trade press, sanctions filings, or specialist databases. \textit{Example: ``What is the primary supplier of potato for McDonald's french fries in the US?''} (supplier identity is not on McDonald's public website.)
\item \textbf{Q4 (Multi-hop, Low-visibility)}: 20 queries; the most challenging category, requiring discovery and connection of multiple low-visibility facts from fragmented sources to construct a supply chain subgraph. \textit{Example: ``List all Procter \& Gamble hair care products that contain both panthenol and biotin, and identify their shared ingredient suppliers.''} (requires tracing product formulations to common upstream suppliers across regulatory filings, supplier press releases, and industry reports.)
\end{itemize}
Each query is accompanied by a ground-truth answer constructed through a two-stage human annotation process. Annotators first compile answers from primary sources (regulatory filings, official product pages, trade press, and supplier directories) along with supporting evidence; domain experts then review every entry for factual correctness, source attribution, and, for Q4, the completeness of the extracted entity-relationship structure.

\section{Experimental Settings}
\label{sec:setup}

\subsection{Dataset and Evaluation Scope}

We evaluate all systems on the SCQA benchmark introduced in \autoref{sec:dataset}. SCQA contains 80 supply chain queries across four quadrants defined by reasoning complexity (single-hop vs.\ multi-hop) and information visibility (high vs.\ low visibility), with 20 queries per quadrant. Each query is paired with a human-verified ground-truth answer; for Q4, we additionally provide ground-truth entity--relationship structures for graph-level evaluation.

\subsection{Compared Systems}

We evaluate \textit{Helicase} against two categories of baselines: frontier LLMs without active web access and agentic frameworks augmented with web search. This comparison separates three capabilities: parametric factual recall, search-enabled reasoning, and uncertainty-guided supply chain knowledge graph construction.

\paragraph{Frontier LLMs (zero-shot).}
Claude Opus 4.6~\citep{anthropic2025claude}, Qwen3-235B-A22B~\citep{qwen2025qwen3}, DeepSeek-V3.2~\citep{deepseek2025v3}, and GLM-5~\citep{glm2024chatglm} are evaluated with direct prompting. These baselines represent the upper bound of parametric knowledge without active web search, testing whether supply chain facts are already recoverable from model training data alone.

\paragraph{Agentic frameworks with web search.}
ReAct~\citep{yao2023react} with a Qwen3-235B backbone is augmented with Serper web search and Jina page reading tools. It implements the foundational reason-act loop for up to 8 steps, but does not include uncertainty guidance or knowledge graph construction. Tree-of-Thoughts (ToT)~\citep{yao2023tot}, also using a Qwen3-235B backbone and identical search tools, extends chain-of-thought by exploring three parallel reasoning branches with self-evaluation and answer merging. Together, these baselines isolate the contribution of \textit{Helicase}'s architectural components: ReAct tests whether a basic search-enabled agentic loop is sufficient, while ToT tests whether structured multi-path deliberation can match iterative uncertainty-guided knowledge graph construction.

\subsection{Implementation}

All systems are evaluated on all 80 SCQA queries under identical conditions: the same web access permissions, the same Serper API search interface, the same Jina Reader page-content extraction tool, the same rate limits, and no query-specific pre-seeding of sources. \textit{Helicase} uses a heterogeneous model configuration: the \textit{Thinking} variant is assigned to the planner, web search, and reasoning agents that drive the helical loop, while the lighter \textit{Instruct} variant is assigned to the coding agent responsible for structured JSON extraction and knowledge graph mutation. All models are served via SiliconFlow. \autoref{tab:agent-models} details the complete agent--model mapping.

\begin{table}[t]
\centering
\caption{Helicase agent roles and their assigned LLM models. Thinking models, which support extended reasoning, are used for planning, web search, and cross-source reasoning; the lighter Instruct variant is used for structured JSON extraction onto the knowledge graph. The same Thinking model also serves as the LLM-based consensus scorer that computes action-layer uncertainty across worker-agent outputs.}
\label{tab:agent-models}
\resizebox{\textwidth}{!}{
\begin{tabular}{@{}llll@{}}
\toprule
Agent Role & Model & Type & Responsibility \\
\midrule
\textbf{Planner} & Qwen3-Next-80B-A3B-Thinking & Thinking & Query decomposition, action generation ($\Phi_{\text{gen}}$), priority scoring, convergence checking \\
\textbf{Web Search Agent} & Qwen3-Next-80B-A3B-Thinking & Thinking & Parallel web search, page reading, citation-tracked answer synthesis \\
\textbf{Reasoning Agent} & Qwen3-Next-80B-A3B-Thinking & Thinking & Cross-source inference over retrieved evidence \\
\textbf{Coding Agent} & Qwen3-Next-80B-A3B-Instruct & Instruct & Structured JSON extraction $\rightarrow$ KG mutations (nodes, edges, decompositions) \\
\bottomrule
\end{tabular}
}
\end{table}

\subsection{Hyperparameters}

All hyperparameters introduced in \autoref{sec:methodology} take the following concrete values in our experiments. \textit{Dynamic query scaling:} parallel search count $n \in [n_{\min}, n_{\max}] = [1, 10]$, triggered when $U(v_{\text{target}}) > \tau_{\text{low}} = 0.3$. \textit{Action planning:} high-uncertainty fact threshold $\tau_{\text{high}} = 0.7$; priority scaling factor $\alpha = 0.3$; redundancy flag threshold $\tau_{\rho} = 0.8$. Per-agent costs in the priority function reflect the relative token and latency overhead of each agent type (web search $>$ reasoning $>$ coding), with exact values provided in our released code. \textit{Convergence:} relative change threshold $\delta_{\text{conv}} = 0.05$ (5\%), patience $P = 3$ consecutive stagnant iterations, hard cap $T_{\max} = 10$ iterations, extended to $T_{\max} = 20$ for Q4 queries. \textit{Graph F1 weights:} $w_E = 0.6$, $w_R = 0.4$. Embeddings use all-MiniLM-L6-v2.

\subsection{Evaluation Protocol}
\label{subsec:evaluation-metrics}

We employ quadrant-specific evaluation metrics that reflect the distinct challenges at each complexity level, along with cross-cutting operational metrics. All metrics use LLM-based semantic matching with ChatGPT 5.5 as judge, allowing semantically equivalent answers to be matched even when surface forms differ, such as ``Pilbara Minerals'' versus ``Pilbara Minerals Ltd.'' 
To guard against bias introduced by the LLM judge, we manually audited a stratified subsample of 20 queries (5 per quadrant) covering all systems, comparing the ChatGPT 5.5 judge's decisions against expert assessment; agreement exceeded 90\% with disagreements concentrated in borderline semantic-equivalence calls. Besides, all ground-truth references, predicted outputs, and judge decisions are released with the SCQA benchmark to enable independent re-evaluation with alternative judges.

\subsection{Quadrant-specific Metrics}
\label{subsec:quadrant-metrics}

We use different metrics for the four SCQA quadrants because each quadrant tests a different capability: direct factual answering, set aggregation, source identification, or graph construction under uncertainty.

\paragraph{Q1: Answer Accuracy (Acc.).}
For single-hop, high-visibility queries, we measure factual correctness by LLM-based semantic equivalence between the predicted answer and the reference answer. For yes/no questions, the boolean answer must match; for entity-based answers, the key factual content must align. Model refusals or ``I do not know'' responses are counted as incorrect because the decision-aid setting requires a usable answer. We additionally report refusal rates in the released artefacts so that abstention can be separated from factual error. All systems are evaluated under a fixed prompting protocol with temperature zero, and prompts are released with the benchmark.

\paragraph{Q2: Set F1 (Precision, Recall, F1).}
For multi-hop, high-visibility queries that require listing multiple items, we use set-based evaluation. The LLM judge first matches semantically equivalent predicted and reference items. Precision is then the fraction of predicted items that are correct, recall is the fraction of reference items recovered, and F1 is their harmonic mean.

\paragraph{Q3: Answer Accuracy + Source Discovery Rate (SDR).}
For single-hop, low-visibility queries, we evaluate both answer correctness and source discovery. Answer accuracy is computed as in Q1. Source Discovery Rate (SDR) measures whether the system identifies relevant and plausible sources for verifying the answer. Given the predicted answer and cited source domains, the LLM judge determines whether the sources are appropriate for validating the supply chain claim. SDR captures whether a system can provide verifiable provenance rather than merely guessing a plausible answer from parametric knowledge.

\paragraph{Q4: Graph F1 (G-F1) + Uncertainty Calibration Error (UCE).}
For multi-hop, low-visibility queries requiring structural reasoning, we evaluate the predicted knowledge graph against the reference graph at both the entity and relation levels. We also assess whether the system's per-fact confidence scores are calibrated.
For entity-level evaluation, a predicted entity is counted as a true positive if the LLM judge confirms that it is semantically equivalent to a reference entity. Let $\hat{V}$ and $V^\star$ denote the predicted and reference entity sets, and let $m_E$ be the number of matched entity pairs. Entity precision and recall are $P_E=m_E/|\hat{V}|$ and $R_E=m_E/|V^\star|$, giving:
\begin{equation}
\text{E-F1}
=
\frac{2P_E R_E}{P_E + R_E}.
\end{equation}

For relation-level evaluation, a predicted triple $\langle h,r,t\rangle$ is counted as a true positive only if both endpoints and the relation type are semantically equivalent to a reference triple. Let $\hat{E}$ and $E^\star$ denote the predicted and reference relation sets, and let $m_R$ be the number of matched relation triples. Relation precision and recall are $P_R=m_R/|\hat{E}|$ and $R_R=m_R/|E^\star|$, giving:
\begin{equation}
\text{R-F1}
=
\frac{2P_R R_R}{P_R + R_R}.
\end{equation}

Graph F1 combines entity-level and relation-level performance:
\begin{equation}
\text{G-F1}
=
w_E \cdot \text{E-F1}
+
w_R \cdot \text{R-F1},
\end{equation}
where $w_E=0.6$ and $w_R=0.4$. Entity discovery is weighted slightly higher because relation recovery depends on correctly identifying the entities involved, and entity identification is often the harder sub-task in low-visibility supply chain queries.
UCE evaluates calibration at the level of individual graph facts. For each evaluated fact $f \in F$, where $f$ is a node or edge in the predicted KG with non-zero evidence, we compute the stated confidence as $c_f = 1-U(f)$, where $U(f)$ is the fact-level uncertainty assigned by the system. The LLM judge then assigns an empirical correctness label $a_f \in \{0,1\}$ by comparing the fact against the reference graph. UCE is the absolute gap between the mean stated confidence and the mean empirical correctness:
\begin{equation}
\text{UCE}
=
\left|
\frac{1}{|F|}\sum_{f \in F} c_f
-
\frac{1}{|F|}\sum_{f \in F} a_f
\right|.
\end{equation}
Lower UCE indicates better calibration: the system's stated confidence is closer to its empirical graph-level accuracy.

\begin{table}[h]
\centering
\caption{Performance on SCQA benchmark. Metrics are defined in \autoref{subsec:evaluation-metrics}. SDR = 0 for systems without web search (no source provenance). Best results in \textbf{bold}.}
\label{tab:main-results}
\resizebox{\textwidth}{!}{
\begin{tabular}{@{}l|c|ccc|cc|cccc@{}}
\toprule
\multirow{2}{*}{System} & Q1 & \multicolumn{3}{c|}{Q2} & \multicolumn{2}{c|}{Q3} & \multicolumn{4}{c}{Q4} \\
 & Acc. & P & R & F1 & Acc. & SDR & E-F1 & R-F1 & G-F1 & UCE$\downarrow$ \\
\midrule
\multicolumn{11}{l}{\cellcolor{gray!10}\textit{Frontier LLMs}} \\
Claude Opus 4.6 & 0.80 & 0.75 & 0.48 & 0.55 & 0.55 & 0.00 & 0.67 & 0.57 & 0.63 & -- \\
GLM-5 & 0.60 & 0.44 & 0.47 & 0.40 & 0.20 & 0.00 & 0.35 & 0.30 & 0.33 & -- \\
DeepSeek-V3.2 & 0.40 & 0.60 & 0.47 & 0.48 & 0.15 & 0.00 & 0.29 & 0.25 & 0.27 & -- \\
Qwen3-235B & 0.30 & 0.52 & 0.44 & 0.44 & 0.05 & 0.00 & 0.26 & 0.22 & 0.25 & -- \\
\midrule
\multicolumn{11}{l}{\cellcolor{gray!10}\textit{Agentic Frameworks}} \\
ReAct (Qwen3-235B) & 0.60 & 0.68 & 0.34 & 0.39 & 0.45 & 0.45 & 0.34 & 0.29 & 0.32 & -- \\
ToT (Qwen3-235B) & 0.55 & 0.50 & 0.28 & 0.30 & 0.35 & 0.35 & 0.41 & 0.35 & 0.39 & -- \\
\midrule
\multicolumn{11}{l}{\cellcolor{gray!10}\textit{Helicase (Ours)}} \\
\textbf{Helicase} & \textbf{0.95} & \textbf{0.90} & \textbf{0.86} & \textbf{0.85} & \textbf{1.00} & \textbf{0.65} & \textbf{0.87} & \textbf{0.83} & \textbf{0.85} & \textbf{0.25} \\
\bottomrule
\end{tabular}
}
\end{table}

\subsection{Ablation Study}
\label{subsec:ablation}

To answer RQ3 and identify which components of Helicase materially drive discovery quality, we conduct an ablation study on the 20 Q4 queries, the hardest quadrant and the one where architectural differences are most informative. We evaluate four ablations: (a) \textit{Helicase without UQ}: remove uncertainty-guided planning and use a uniform action-priority scheme; (b) \textit{Helicase without multiplicative accumulation}: replace multiplicative accumulation of evidence with a minimum-based rule; (c) \textit{Helicase without dynamic parallel search}: disable dynamic parallel query scaling and fix $n=1$; (d) \textit{Helicase without MAS}: replace the multi-agent pipeline with a single LLM agent performing search, extraction, and graph construction in one loop. Results are reported in \autoref{tab:ablation}. Uncertainty-guided planning drives discovery quality (G-F1 $-$0.12) and is a prerequisite for calibrated UCE. Multiplicative accumulation is the dominant calibration mechanism: the min-rule stagnates at the first observation, inflating UCE to 0.41. Multi-agent specialisation is the single largest driver of end-to-end performance, with a one-agent loop falling below all agentic baselines in \autoref{tab:main-results}. Dynamic parallel search contributes more modestly but remains meaningful for low-visibility queries where individual searches are noisy. A natural follow-up question is whether the KG-construction and uncertainty-scoring capabilities could be \textit{retrofitted} onto a baseline such as ReAct via a post-hoc extraction and calibration pass. Our Helicase without MAS ablation directly addresses this: a one-agent loop performs search, extraction, and graph construction in a single trajectory, producing a structured KG as output. Its G-F1 collapses to 0.45, well below Helicase's 0.85 and only marginally above the agentic baselines (ReAct 0.32, ToT 0.39) despite producing a structured KG. This indicates that the value of Helicase's KG does not arise from the act of post-hoc extraction but from the tight coupling between uncertainty-guided search, incremental graph construction, and multiplicative evidence accumulation across iterations. Similarly, without multi-layer uncertainty tracking inside the loop, a post-hoc calibration layer would have only final-answer distributions to work with, not the evidence-level signals needed to assign per-fact confidence.

\begin{table}[h]
\centering
\caption{Ablation study on Q4 (Multi-hop, Low-visibility). Removing any core component substantially degrades both graph quality and uncertainty calibration. UCE ``--'' indicates the variant cannot produce calibrated uncertainty estimates.}
\label{tab:ablation}
\begin{tabular}{@{}lcccc@{}}
\toprule
Variant & E-F1 & R-F1 & G-F1 & UCE$\downarrow$ \\
\midrule
\textbf{Helicase (full)} & \textbf{0.87} & \textbf{0.83} & \textbf{0.85} & \textbf{0.25} \\
\midrule
without UQ (uniform planning) & 0.76 & 0.71 & 0.73 & -- \\
without multiplicative accumulation (min-based update) & 0.69 & 0.62 & 0.65 & 0.41 \\
without dynamic parallel search ($n=1$) & 0.71 & 0.66 & 0.68 & 0.34 \\
without MAS (one-agent loop) & 0.49 & 0.41 & 0.45 & -- \\
\bottomrule
\end{tabular}
\end{table}

\subsection{Computational Cost}

\begin{table}[t]
\centering
\footnotesize
\setlength{\tabcolsep}{20pt}
\caption{Estimated computational cost and output comparison per query. Costs are estimated from observed token usage, tool calls, and provider list prices at the time of experimentation. Helicase costs vary with query depth because the number of helical iterations and parallel evidence-gathering actions is adaptive.}
\label{tab:efficiency}
\begin{tabular}{@{}lcccc@{}}
\toprule
System & Iters & \$/query & KG & UQ \\
\midrule
Claude Opus 4.6 & 1.0 & 0.45 & \ding{55} & \ding{55} \\
ReAct (Qwen3-235B) & 4.0 & 0.15 & \ding{55} & \ding{55} \\
ToT (Qwen3-235B) & 3.0 & 0.20 & \ding{55} & \ding{55} \\
\midrule
Helicase (heterogeneous Qwen3) & 5.0 & 0.20--0.45 & \ding{51} & \ding{51} \\
\bottomrule
\end{tabular}
\end{table}

\autoref{tab:efficiency} reports the estimated computational cost per query. Costs are estimated from observed token usage, tool calls, and provider list prices at the time of experimentation. Helicase costs approximately \$0.20--\$0.45 per query, depending on query complexity and the number of helical iterations required. This places Helicase within the same broad cost range as the strongest LLM and agentic baselines, while producing a qualitatively different output: a structured supply chain knowledge graph with per-fact uncertainty scores rather than an unstructured textual answer.
The cost variation reflects Helicase's adaptive execution policy. Simpler queries require fewer evidence-gathering and reasoning actions, whereas multi-hop, low-visibility queries require additional search, verification, and graph-construction steps. In a typical multi-iteration run, the system performs multiple planner, search, reasoning, and coding-agent calls, with most of the cost arising from reasoning-intensive agents. Full token logs and cost-estimation scripts will be released with the experimental artefacts to support reproducibility.

Beyond API cost, the more relevant comparison is with the manual supplier-mapping workflow that Helicase is designed to augment. Prior DSCS research notes that traditional approaches such as manual mapping, supplier monitoring, and third-party auditing tend to be costly and require substantial time and effort \citep{brintrup2024digital}. In this context, Helicase's sub-dollar per-query cost suggests that uncertainty-guided supply chain knowledge graph construction can serve as a low-cost decision-aid layer for preliminary supply chain mapping and evidence triage. We do not interpret this as replacing expert validation; rather, Helicase can reduce the amount of analyst effort spent on initial evidence gathering, graph assembly, and uncertainty prioritisation.
\section{Qualitative Query-Level Results}
\label{sec:qualitative-results}

The aggregate results in the previous section show that \textit{Helicase} outperforms baseline systems across the SCQA quadrants. To make these results interpretable at the level of individual queries, we present two representative query-level examples. The first illustrates a multi-hop, low-visibility automotive supply chain query requiring cross-tier reasoning from raw materials to downstream Tesla products. The second illustrates a consumer-goods formulation query requiring the integration of product ingredients, suppliers, and intermediary distributors. These examples are not intended as standalone case studies; rather, they provide detailed traces of how \textit{Helicase} constructs supply chain knowledge graphs and assigns uncertainty to entities and relationships.

\subsection{Representative Query Trace: Tesla Lithium Supply Chain (Q64)}

To illustrate \textit{Helicase}'s behaviour on a multi-hop, low-visibility query, we examine Q64: \textit{``Which Tesla components use lithium from Australian mines?''} This query requires tracing lithium from ore extraction in Western Australia through refining, battery cell manufacturing, and vehicle or energy-storage assembly. The relevant evidence is fragmented across mining offtake agreements, corporate filings, battery supplier announcements, and automotive industry analyses.

\begin{figure}[h]
    \centering
    \includegraphics[width=\textwidth]{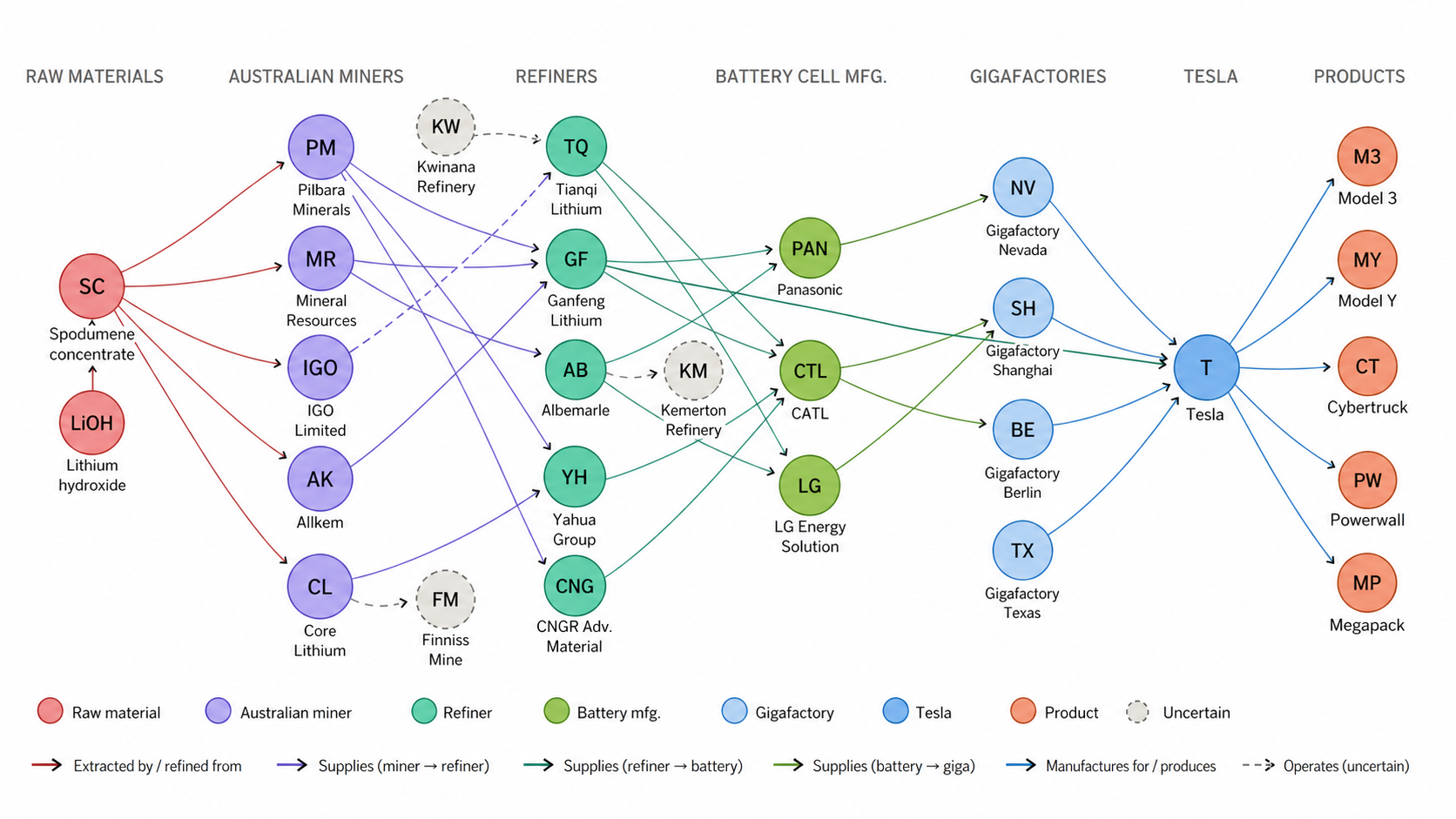}
    \caption{Knowledge graph constructed by \textit{Helicase} for Q64: \textit{``Which Tesla components use lithium from Australian mines?''} The graph contains 28 nodes and 45 edges spanning seven tiers: raw minerals, Australian mining companies, lithium refiners, battery cell manufacturers, Gigafactories, Tesla as the OEM hub, and downstream Tesla products. Grey dashed nodes denote high-uncertainty entities, such as facilities with unconfirmed operational status. The cross-tier edge from Ganfeng Lithium directly to Tesla represents a reported direct supply relationship that bypasses the battery cell manufacturer tier. The system completed five helical iterations and converged at $U_{\text{memory}} = 0.20$.}
    \label{fig:case-study-kg}
\end{figure}

\paragraph{Iterative graph construction.}
\textit{Helicase} completed five helical iterations and constructed a seven-tier supply chain knowledge graph with 28 nodes and 45 edges, as shown in \autoref{fig:case-study-kg}. Of the 28 nodes, 25 are confirmed entities organised across seven tiers (enumerated below), and the remaining 3 are high-uncertainty entities flagged by the system (shown as grey dashed nodes in \autoref{fig:case-study-kg}) representing facilities or relationships with unconfirmed operational status. The 25 confirmed entities are:
\begin{enumerate}
    \item Two raw materials: spodumene concentrate and lithium hydroxide.
    \item Five Australian mining companies: Pilbara Minerals, Mineral Resources, IGO Limited, Allkem, and Core Lithium.
    \item Five lithium refiners and processors: Tianqi Lithium, Ganfeng Lithium, Albemarle, Yahua Group, and CNGR Advanced Material.
    \item Three battery cell manufacturers: Panasonic, CATL, and LG Energy Solution.
    \item Four Gigafactories: Nevada, Shanghai, Berlin, and Texas.
    \item Tesla as the central OEM.
    \item Five downstream products: Model~3, Model~Y, Cybertruck, Powerwall, and Megapack.
\end{enumerate}
The memory-layer uncertainty decreased from 1.00 to 0.20 over the five iterations. Upstream mining--refiner links retained higher uncertainty ($U > 0.30$), reflecting less direct or more fragmented evidence, while downstream factory--product links showed lower uncertainty ($U < 0.10$) due to stronger public documentation.

\paragraph{Query-level findings.}
The resulting graph shows how Australian lithium can flow through multiple intermediate tiers before reaching Tesla products. At the mining level, \textit{Helicase} identified links from Pilbara Minerals to several Chinese refiners, including Ganfeng Lithium, Yahua Group, and CNGR Advanced Material. It also represented IGO Limited's joint venture relationship with Tianqi Lithium, highlighting Australia's upstream role in lithium supply chains connected to Chinese midstream refining capacity. At the refining level, the graph includes a reported direct supply relationship between Ganfeng Lithium and Tesla, which creates a cross-tier link not captured by a simple linear mine--refiner--cell manufacturer hierarchy. At the manufacturing level, the graph maps supplier--factory relationships involving Panasonic, CATL, and LG Energy Solution across Tesla Gigafactories.

\paragraph{Baseline comparison.}
The baseline systems recovered only partial fragments of this structure. Claude Opus 4.6 identified that Tesla sources lithium from Australia and named some upstream firms, but did not reconstruct the refiner tier or Gigafactory-level allocation. ReAct retrieved relevant mining-industry information but did not trace material flow through refiners to battery cell manufacturers and Tesla products. In contrast, \textit{Helicase} assembled a multi-tier graph from mine to downstream product, while assigning uncertainty to individual entities and relationships.

\subsection{Representative Query Trace: Procter \& Gamble Hair Care Formulations (Q17)}

We next examine Q17: \textit{``List all Procter \& Gamble hair care products that contain both panthenol and biotin, and identify their shared ingredient suppliers.''} This query differs from the Tesla lithium example because it requires cross-referencing product formulation information with ingredient-supplier evidence. Rather than a mostly linear mine-to-product chain, the resulting structure is closer to a many-to-many consumer-goods supply network, where multiple products may share ingredients, suppliers, and intermediary distributors.

\paragraph{Iterative graph construction.}
\textit{Helicase} completed four helical iterations and converged at $U_{\text{memory}} = 0.20$. The system produced a bipartite supply chain knowledge graph with 17 nodes and 29 edges, including eight P\&G hair care products, two active ingredients, five suppliers, and two intermediary distributors. The graph linked panthenol-containing products across P\&G's Pantene, Herbal Essences, and Head \& Shoulders lines to suppliers such as DSM-Firmenich and BASF, while biotin sourcing was associated with suppliers including Lonza and Zhejiang NHU Co.

\paragraph{Query-level findings.}
The constructed graph highlights shared ingredient dependencies across multiple P\&G product lines. It also surfaces a cross-category dependency pattern: suppliers associated with panthenol may also serve pharmaceutical, cosmetics, or dietary-supplement markets. This kind of dependency is difficult to observe from a single product ingredient list, because the relevant evidence is distributed across formulation records, supplier directories, and industry sources. The uncertainty annotations allow high-confidence product--ingredient links to be separated from more tentative upstream supplier or cross-industry dependency links.

\paragraph{Baseline comparison.}
The baselines struggled to connect product formulation evidence to upstream supplier relationships. Claude Opus correctly listed some relevant products but returned unsupported or incorrect supplier names, reflecting the risk of hallucination in low-visibility commercial relationships. ReAct and ToT retrieved ingredient disclosure pages but did not reliably connect those ingredients to upstream suppliers. \textit{Helicase} achieved Graph F1 = 0.85 with UCE = 0.25, showing that its performance on the Tesla lithium query is not specific to automotive supply chains, but extends to many-to-many consumer-goods supply networks.

\section{Discussion and Managerial Implications}
\label{sec:discussion}

Our results (addressing RQ1 and RQ2) show that an agentic LLM-based multi-agent system can autonomously answer multi-hop, low-visibility supply chain queries while producing calibrated confidence in its outputs. The ablation (RQ3) identifies the uncertainty-guided planning, multiplicative evidence accumulation, and multi-agent specialisation as the load-bearing components. We now discuss the implications for supply chain practice and for the research agenda on agentic supply chain management.

\paragraph{Adaptive, uncertainty-driven discovery for supply chain MAS}
Existing multi-agent systems for supply chain applications rely on static prompts and fixed architectures: the same agent set, the same templates, and the same iteration budget regardless of query complexity \citep{xu2021bots}. Helicase provides a framework in which the MAS \emph{architects itself per query}, with the planner deciding (i) which prompt to issue to each agent, (ii) which direction in the knowledge graph to discover next, (iii) how many agents to call in parallel, and (iv) how many helical iterations to run. Resource allocation thus tracks the actual difficulty of each query rather than a precommitted budget, with termination governed by sustained stagnation in $U_{\text{memory}}$. This makes Helicase practically deployable on high-stakes supply chain problems (multi-tier supplier discovery, ingredient and material traceability, cross-industry dependency mapping, and disruption-risk assessment), with computational cost that scales \emph{adaptively} with query difficulty rather than uniformly across queries. To our knowledge, Helicase is the first autonomous, self-thinking multi-agent system for supply chain applications, able to decide its own agent prompts, search directions, and iteration depth without hand-crafted workflows.

\paragraph{On-demand visibility for supply chain managers}
Traditional supply chain visibility studies require weeks of analyst time, access to proprietary databases, and expert judgement to assemble multi-tier supplier maps. Helicase reduces this cycle to minutes for any product whose supply chain is documented in public sources. Procurement, risk, and sustainability managers can pose natural-language questions such as \textit{``Which of our Tier-2 cobalt suppliers ultimately source from Democratic Republic of the Congo?''} and receive a structured, evidence-linked answer without hiring consultants or subscribing to commercial databases. This directly addresses the accessibility barrier that has limited visibility research to large firms with dedicated analytics teams.

\paragraph{Trustworthy agentic decisions via calibrated uncertainty}
Recent agentic-SCM work has warned that LLM agents produce confidently wrong decisions \citep{menache2025generative, chen2025manager} and inherit systematic biases from pre-training. Helicase's three-layer uncertainty quantification mitigates this failure mode at the system level: each fact in the output KG carries a calibrated confidence score (UCE = 0.25), so downstream users, whether human managers or downstream agents, can weight evidence accordingly rather than treating all agent outputs as uniformly reliable. For decision-making under uncertainty, this is the difference between an agentic system that is \textit{useful} and one that is \textit{actionable} in regulated or high-stakes procurement contexts.

\paragraph{Empirical substrate for agentic-SCM research}
Beyond its direct managerial use, Helicase's output is a structured, uncertainty-annotated knowledge graph that can be consumed by downstream SCM research pipelines such as supplier network studies, sustainability audits, risk scoring, and disruption analysis. Rather than relying on survey-based or proprietary datasets that are expensive, incomplete, or outdated, researchers can use agentic discovery as an upstream data collection layer. The SCQA benchmark further provides a common yardstick for evaluating future agentic SCM systems across a controlled complexity spectrum.

\paragraph{Illustrative decision scenario}
To make the decision-aid framing concrete, consider the P\&G panthenol/biotin scenario from our Q17 case study (§6.4). A procurement manager at P\&G hair care receives an internal escalation: a critical-materials audit requires identifying cross-category single-supplier dependencies. Using Helicase, the manager issues the Q17 query and receives, within minutes, a 17-node knowledge graph showing that (i) panthenol across three brand families (Pantene, Herbal Essences, Head \& Shoulders) concentrates on two suppliers, DSM-Firmenich and BASF, each carrying $U < 0.2$ indicating high confidence; (ii) biotin concentrates on Lonza and Zhejiang NHU with slightly higher uncertainty ($U \approx 0.3$); and (iii) DSM-Firmenich additionally serves pharmaceutical and cosmetics clients, evidence for which is at $U \approx 0.35$. With this artefact, the manager can immediately act on the high-confidence facts (initiate dual-sourcing discussions for panthenol; flag the Pantene--Herbal Essences co-dependency for risk review) while routing the cross-industry dependency finding to a human analyst for verification given its moderate uncertainty. The decision that changes because of Helicase is not the analyst's final recommendation but the allocation of analyst time: scarce expert attention is directed at the small set of genuinely uncertain facts, not at reconstructing supplier maps from scratch. This is the specific sense in which Helicase operates as an agentic decision aid.

\section{Limitations}

Our study has several limitations. \textit{First}, our ``low-visibility'' queries target information that exists in public web sources but is difficult to find; truly \textit{invisible} supply chain information (proprietary contracts, trade-secret product formulations, classified agreements, and informal relationships) cannot be discovered through web search and remains outside the system's reach. \textit{Second}, the system requires web access and LLM API calls, making it subject to service availability, rate limits, and the temporal validity of web sources. \textit{Third}, the LLM-based consensus mechanism assumes that agreement across independent web sources signals factual truth. For supply chain topics where a single erroneous claim is recycled across analyst reports, trade press, and aggregator sites, the multiplicative accumulation update will assign that claim high confidence. The LLM-based consensus scoring mitigates but does not eliminate this failure mode; detecting source dependence (e.g., that two apparently independent pages in fact cite each other) remains an open problem for agentic web-grounded systems and a natural extension of the uncertainty framework presented here. \textit{Fourth}, the system is vulnerable to adversarial information attacks: coordinated misinformation campaigns, SEO-poisoned content, or knowledge-graph-poisoning attacks that seed false claims across multiple seemingly independent web sources can pass the LLM consensus check and be accumulated as high-confidence facts. Defending against such attacks would require provenance-aware source authentication and cross-modal verification beyond textual consensus, both of which lie outside the current framework. \textit{Finally}, several complex supply chain analytical problems, such as temporal and longitudinal analysis of how supplier relationships evolve over time, disruption-event reasoning, and cross-organisational data-sharing scenarios, were outside our evaluation scope. These analyses can naturally be carried out on top of the supply chain knowledge graphs generated by Helicase, which already encode entities, relations, and per-fact uncertainty in a structured form amenable to graph-analytic, time-series, and event-based extensions, and we view them as our future work.

\section{Conclusion and Future Work}
\label{sec:conclusion}

We presented Helicase, an agentic LLM-based multi-agent system for uncertainty-aware supply chain discovery. Helicase moves beyond narrative synthesis to autonomous knowledge graph construction: through a helical process of query enrichment, specialised multi-agent execution, and three-layer uncertainty quantification, it progressively assembles supply chain graphs annotated with calibrated confidence estimates.
Our evaluation on the SCQA benchmark directly answers the three research questions posed in the Introduction. For \textbf{RQ1}, Helicase successfully answers multi-hop, low-visibility supply chain queries that defeat frontier LLMs and agentic search baselines, achieving 85\% Graph F1 on the hardest quadrant, which no baseline matches. For \textbf{RQ2}, the three-layer uncertainty framework delivers calibrated confidence (UCE = 0.25), a capability absent from all baselines tested, and a direct response to the trust and verification concerns raised by recent agentic-SCM research. For \textbf{RQ3}, the ablation study identifies uncertainty-guided planning, multiplicative evidence accumulation, and multi-agent specialisation as the load-bearing architectural components, with dynamic parallel search contributing more modestly.

For the production research community, this work offers three takeaways. First, agentic LLMs can deliver autonomous supply chain discovery at a quality previously requiring weeks of expert analyst work, addressing the scalability and coordination failures that stalled classical MAS in SCM. Second, explicit uncertainty quantification is not optional: without it, agentic systems produce confidently wrong outputs that are not actionable in managerial decision-making. Third, SCQA provides a common evaluation substrate for future agentic-SCM research. Future work will extend SCQA to additional industrial domains and pursue several natural extensions that consume or build on the agentic discovery substrate introduced here: (i) downstream network analytics on the constructed knowledge graphs, including ripple-effect and disruption-propagation modelling; (ii) human-in-the-loop verification workflows that route high-uncertainty facts ($U > \tau_{\text{high}}$) to expert review; (iii) negotiation and inter-organisational coordination between discovery agents for cross-firm supply chain mapping; (iv) temporal discovery, tracking how supplier relationships, formulations, and corporate ownership evolve over time; and (v) integration with enterprise data sources to reach supply chain regions beyond the public web.

\section*{Data Availability Statement}

The dataset supporting this study will be made openly available upon publication. The SCQA benchmark, including all 80 supply chain queries, ground-truth answers, supporting evidence, annotation guidelines, predicted outputs, and judge decisions reported in this paper, will be released under a permissive open licence to enable independent re-evaluation and future extension. To the best of our knowledge, SCQA is the first public benchmark specifically designed to evaluate deep-research and agentic LLM systems for supply chain structural inference, covering both single-hop and multi-hop queries under high- and low-visibility information settings. All source materials used to construct and evaluate SCQA are publicly accessible; no proprietary, confidential, or restricted-access data were used in this study.

\bibliographystyle{iclr2024_conference}
\bibliography{references}

@article{brintrup2024digital,
  title={Digital supply chain surveillance using artificial intelligence: definitions, opportunities and risks},
  author={Brintrup, Alexandra and Kosasih, Edward and Schaffer, Philipp and Zheng, Ge and Demirel, Guven and MacCarthy, Bart L},
  journal={International Journal of Production Research},
  volume={62},
  number={13},
  pages={4674--4695},
  year={2024},
  publisher={Taylor \& Francis}
}

@article{kosasih2022machine,
  title={A machine learning approach for predicting hidden links in supply chain with graph neural networks},
  author={Kosasih, Edward Elson and Brintrup, Alexandra},
  journal={International Journal of Production Research},
  volume={60},
  number={17},
  pages={5380--5393},
  year={2022},
  publisher={Taylor \& Francis}
}

@article{jannelli2025agentic,
  title={Agentic LLMs in the supply chain: towards autonomous multi-agent consensus-seeking},
  author={Jannelli, Valeria and Schoepf, Stefan and Bickel, Matthias and Netland, Torbj{\o}rn and Brintrup, Alexandra},
  journal={International Journal of Production Research},
  pages={1--31},
  year={2025},
  publisher={Taylor \& Francis}
}

@article{quan2024invagent,
  title={Invagent: A large language model based multi-agent system for inventory management in supply chains},
  author={Quan, Yinzhu and Liu, Zefang},
  journal={arXiv preprint arXiv:2407.11384},
  year={2024}
}

@inproceedings{wu2025agentic,
  title={Agentic reasoning: A streamlined framework for enhancing llm reasoning with agentic tools},
  author={Wu, Junde and Zhu, Jiayuan and Liu, Yuyuan and Xu, Min and Jin, Yueming},
  booktitle={Proceedings of the 63rd Annual Meeting of the Association for Computational Linguistics (Volume 1: Long Papers)},
  pages={28489--28503},
  year={2025}
}

@article{zhang2025deep,
  title={Deep research: A survey of autonomous research agents},
  author={Zhang, Wenlin and Li, Xiaopeng and Zhang, Yingyi and Jia, Pengyue and Wang, Yichao and Guo, Huifeng and Liu, Yong and Zhao, Xiangyu},
  journal={arXiv preprint arXiv:2508.12752},
  year={2025}
}

@article{dong2025safesearch,
  title={SafeSearch: Automated Red-Teaming for the Safety of LLM-Based Search Agents},
  author={Dong, Jianshuo and Guo, Sheng and Wang, Hao and Chen, Xun and Liu, Zhuotao and Zhang, Tianwei and Xu, Ke and Huang, Minlie and Qiu, Han},
  journal={arXiv preprint arXiv:2509.23694},
  year={2025}
}

@inproceedings{zhao2025uncertainty,
  title={Uncertainty propagation on llm agent},
  author={Zhao, Qiwei and Li, Dong and Liu, Yanchi and Cheng, Wei and Sun, Yiyou and Oishi, Mika and Osaki, Takao and Matsuda, Katsushi and Yao, Huaxiu and Zhao, Chen and others},
  booktitle={Proceedings of the 63rd Annual Meeting of the Association for Computational Linguistics (Volume 1: Long Papers)},
  pages={6064--6073},
  year={2025}
}

@article{zhao2024saup,
  title={SAUP: Situation Awareness Uncertainty Propagation on LLM Agent},
  author={Zhao, Qiwei and Zhao, Xujiang and Liu, Yanchi and Cheng, Wei and Sun, Yiyou and Oishi, Mika and Osaki, Takao and Matsuda, Katsushi and Yao, Huaxiu and Chen, Haifeng},
  journal={arXiv preprint arXiv:2412.01033},
  year={2024}
}

@article{han2024towards,
  title={Towards uncertainty-aware language agent},
  author={Han, Jiuzhou and Buntine, Wray and Shareghi, Ehsan},
  journal={arXiv preprint arXiv:2401.14016},
  year={2024}
}

@article{kirchhof2025position,
  title={Position: Uncertainty quantification needs reassessment for large-language model agents},
  author={Kirchhof, Michael and Kasneci, Gjergji and Kasneci, Enkelejda},
  journal={arXiv preprint arXiv:2505.22655},
  year={2025}
}

@article{xu2024multi,
  title={Multi-agent systems and foundation models enable autonomous supply chains: Opportunities and challenges},
  author={Xu, Liming and Almahri, Sara and Mak, Stephen and Brintrup, Alexandra},
  journal={IFAC-PapersOnLine},
  volume={58},
  number={19},
  pages={795--800},
  year={2024},
  publisher={Elsevier}
}

@article{huang2025deep,
  title={Deep research agents: A systematic examination and roadmap},
  author={Huang, Yuxuan and Chen, Yihang and Zhang, Haozheng and Li, Kang and Zhou, Huichi and Fang, Meng and Yang, Linyi and Li, Xiaoguang and Shang, Lifeng and Xu, Songcen and others},
  journal={arXiv preprint arXiv:2506.18096},
  year={2025}
}

@article{nahar2025catch,
  title={Catch Me if You Search: When Contextual Web Search Results Affect the Detection of Hallucinations},
  author={Nahar, Mahjabin and Lee, Eun-Ju and Park, Jin Won and Lee, Dongwon},
  journal={arXiv preprint arXiv:2504.01153},
  year={2025}
}

@article{simangunsong2012supply,
  title={Supply-chain uncertainty: a review and theoretical foundation for future research},
  author={Simangunsong, Eliot and Hendry, Linda C and Stevenson, Mark},
  journal={International journal of production research},
  volume={50},
  number={16},
  pages={4493--4523},
  year={2012},
  publisher={Taylor \& Francis}
}

@article{tseng2018decision,
  title={Decision-making model for sustainable supply chain finance under uncertainties},
  author={Tseng, Ming-Lang and Wu, Kuo-Jui and Hu, Jiayao and Wang, Chin-Hsin},
  journal={International Journal of Production Economics},
  volume={205},
  pages={30--36},
  year={2018},
  publisher={Elsevier}
}

@inproceedings{chan2024visibility,
  title={Visibility into AI agents},
  author={Chan, Alan and Ezell, Carson and Kaufmann, Max and Wei, Kevin and Hammond, Lewis and Bradley, Herbie and Bluemke, Emma and Rajkumar, Nitarshan and Krueger, David and Kolt, Noam and others},
  booktitle={Proceedings of the 2024 ACM Conference on Fairness, Accountability, and Transparency},
  pages={958--973},
  year={2024}
}

@article{kosasih2024review,
  title={A review of explainable artificial intelligence in supply chain management using neurosymbolic approaches},
  author={Kosasih, Edward Elson and Papadakis, Emmanuel and Baryannis, George and Brintrup, Alexandra},
  journal={International Journal of Production Research},
  volume={62},
  number={4},
  pages={1510--1540},
  year={2024},
  publisher={Taylor \& Francis}
}

@inproceedings{zhou2025efficient,
  title={Efficient multi-agent collaboration with tool use for online planning in complex table question answering},
  author={Zhou, Wei and Mesgar, Mohsen and Friedrich, Annemarie and Adel, Heike},
  booktitle={Findings of the Association for Computational Linguistics: NAACL 2025},
  pages={945--968},
  year={2025}
}

@article{agrawal2024supply,
  title={Supply chain visibility: A Delphi study on managerial perspectives and priorities},
  author={Agrawal, Tarun Kumar and Kalaiarasan, Ravi and Olhager, Jan and Wiktorsson, Magnus},
  journal={International Journal of Production Research},
  volume={62},
  number={8},
  pages={2927--2942},
  year={2024},
  publisher={Taylor \& Francis}
}

@article{kalaiarasan2023supply,
  title={Supply chain visibility for improving inbound logistics: a design science approach},
  author={Kalaiarasan, Ravi and Agrawal, Tarun Kumar and Olhager, Jan and Wiktorsson, Magnus and Hauge, Jannicke Baalsrud},
  journal={International Journal of Production Research},
  volume={61},
  number={15},
  pages={5228--5243},
  year={2023},
  publisher={Taylor \& Francis}
}

@article{zheng2025enhancing,
  title={Enhancing supply chain visibility with generative AI: an exploratory case study on relationship prediction in knowledge graphs},
  author={Zheng, Ge and Brintrup, Alexandra},
  journal={International Journal of Production Research},
  pages={1--23},
  year={2025},
  publisher={Taylor \& Francis}
}

@article{kosasih2025towards,
  title={Towards trustworthy AI for link prediction in supply chain knowledge graph: a neurosymbolic reasoning approach},
  author={Kosasih, Edward Elson and Brintrup, Alexandra},
  journal={International Journal of Production Research},
  volume={63},
  number={6},
  pages={2268--2290},
  year={2025},
  publisher={Taylor \& Francis}
}

@article{wichmann2020extracting,
  title={Extracting supply chain maps from news articles using deep neural networks},
  author={Wichmann, Pascal and Brintrup, Alexandra and Baker, Simon and Woodall, Philip and McFarlane, Duncan},
  journal={International Journal of Production Research},
  volume={58},
  number={17},
  pages={5320--5336},
  year={2020},
  publisher={Taylor \& Francis}
}

@misc{ivanov2016supply,
  title={Supply chain dynamics, control and disruption management},
  author={Ivanov, Dmitry and Mason, Scott J and Hartl, Richard},
  journal={International Journal of Production Research},
  volume={54},
  number={1},
  pages={1--7},
  year={2016},
  publisher={Taylor \& Francis}
}

@misc{tiwari2024supply,
  title={Supply chain digitisation and management},
  author={Tiwari, Manoj Kumar and Bidanda, Bopaya and Geunes, Joseph and Fernandes, Kiran and Dolgui, Alexandre},
  journal={International Journal of Production Research},
  volume={62},
  number={8},
  pages={2918--2926},
  year={2024},
  publisher={Taylor \& Francis}
}

@article{kosasih2024towards,
  title={Towards knowledge graph reasoning for supply chain risk management using graph neural networks},
  author={Kosasih, Edward Elson and Margaroli, Fabrizio and Gelli, Simone and Aziz, Ajmal and Wildgoose, Nick and Brintrup, Alexandra},
  journal={International Journal of Production Research},
  volume={62},
  number={15},
  pages={5596--5612},
  year={2024},
  publisher={Taylor \& Francis}
}

@article{chen2019understanding,
  title={Understanding dataset design choices for multi-hop reasoning},
  author={Chen, Jifan and Durrett, Greg},
  journal={arXiv preprint arXiv:1904.12106},
  year={2019}
}

@article{aziz2021data,
  title={Data considerations in graph representation learning for supply chain networks},
  author={Aziz, Ajmal and Kosasih, Edward Elson and Griffiths, Ryan-Rhys and Brintrup, Alexandra},
  journal={arXiv preprint arXiv:2107.10609},
  year={2021}
}

@misc{anthropic2025claude,
  title={Claude {Opus} 4 and {Claude} {Sonnet} 4 {System} {Card}},
  author={{Anthropic}},
  year={2025},
  howpublished={\url{https://www.anthropic.com/news/claude-4}}
}

@inproceedings{yao2023react,
  title={ReAct: Synergizing Reasoning and Acting in Language Models},
  author={Yao, Shunyu and Zhao, Jeffrey and Yu, Dian and Du, Nan and Shafran, Izhak and Narasimhan, Karthik and Cao, Yuan},
  booktitle={International Conference on Learning Representations},
  year={2023}
}

@inproceedings{yao2023tot,
  title={Tree of Thoughts: Deliberate Problem Solving with Large Language Models},
  author={Yao, Shunyu and Yu, Dian and Zhao, Jeffrey and Shafran, Izhak and Griffiths, Thomas L. and Cao, Yuan and Narasimhan, Karthik},
  booktitle={Advances in Neural Information Processing Systems},
  year={2023}
}

@article{deepseek2025v3,
  title={DeepSeek-V3 technical report},
  author={{DeepSeek-AI}},
  journal={arXiv preprint arXiv:2412.19437},
  year={2024}
}

@article{qwen2025qwen3,
  title={Qwen3 technical report},
  author={{Qwen Team}},
  journal={arXiv preprint arXiv:2505.09388},
  year={2025}
}

@article{glm2024chatglm,
  title={{ChatGLM}: A family of large language models from {GLM-130B} to {GLM-4} all tools},
  author={Zeng, Aohan and Liu, Xiao and Du, Zhengxiao and Wang, Zihan and Lai, Hanyu and Ding, Ming and Yang, Zhuoyi and Xu, Yifan and Zheng, Wendi and Xia, Xiao and others},
  journal={arXiv preprint arXiv:2406.12793},
  year={2024}
}

@article{xu2021bots,
  title={Will bots take over the supply chain? Revisiting agent-based supply chain automation},
  author={Xu, Liming and Mak, Stephen and Brintrup, Alexandra},
  journal={International Journal of Production Economics},
  volume={241},
  pages={108279},
  year={2021},
  publisher={Elsevier}
}

@article{choi2001supply,
  title={Supply networks and complex adaptive systems: control versus emergence},
  author={Choi, Thomas Y and Dooley, Kevin J and Rungtusanatham, Manus},
  journal={Journal of Operations Management},
  volume={19},
  number={3},
  pages={351--366},
  year={2001},
  publisher={Elsevier}
}

@article{bubeck2023sparks,
  title={Sparks of artificial general intelligence: Early experiments with {GPT}-4},
  author={Bubeck, S{\'e}bastien and Chandrasekaran, Varun and Eldan, Ronen and Gehrke, Johannes and Horvitz, Eric and Kamar, Ece and Lee, Peter and others},
  journal={arXiv preprint arXiv:2303.12712},
  year={2023}
}

@article{li2025chatsync,
  title={{ChatSync}: Large language model enabled spatial-temporal knowledge reasoning for production logistics synchronization},
  author={Li, Junkai and Zhao, Zhiheng and Yang, Chen and Huang, Shan and Lee, Loo Hay and Huang, George Q},
  journal={IEEE Internet of Things Journal},
  year={2025},
  publisher={IEEE}
}

@article{menache2025generative,
  title={How generative {AI} improves supply chain management},
  author={Menache, Ishai and Pathuri, Jayanth and Simchi-Levi, David and Linton, Tom},
  journal={Harvard Business Review},
  volume={103},
  number={1--2},
  pages={86--95},
  year={2025}
}

@article{chen2025manager,
  title={A manager and an {AI} walk into a bar: does {ChatGPT} make biased decisions like we do?},
  author={Chen, Yang and Kirshner, Samuel N and Ovchinnikov, Anton and Andiappan, Meena and Jenkin, Tracy},
  journal={Manufacturing \& Service Operations Management},
  year={2025}
}

@article{brintrup2018predicting,
  title={Predicting Hidden Links in Supply Networks},
  author={Brintrup, A. and Wichmann, P. and Woodall, P. and McFarlane, D. and Nicks, E. and Krechel, W.},
  journal={Complexity},
  volume={2018},
  number={1},
  pages={9104387},
  year={2018},
  doi={10.1155/2018/9104387},
  publisher={Wiley}
}

\end{document}